\documentclass{article}

\usepackage[final]{neurips_2023}

\usepackage[utf8]{inputenc} 
\usepackage[T1]{fontenc}    
\usepackage{hyperref}       
\usepackage{url}            
\usepackage{booktabs}       
\usepackage{amsfonts}       
\usepackage{nicefrac}       
\usepackage{microtype}      
\usepackage{xcolor}         
\usepackage{graphicx}       
\usepackage{tikz}
\usepackage[many]{tcolorbox}
\usepackage{textcomp}
\usepackage{multirow}
\usepackage{wrapfig}
\usepackage{caption}
\usepackage{amssymb}
\usepackage{algorithm, algorithmic}

\def\Snospace~{\S{}}

\newcommand{\bR}{\mathbb{R}}
\newcommand{\bN}{\mathbb{N}}
\newcommand{\mc}[1]{\mathcal{#1}}
\newcommand{\ra}{\rightarrow}

\definecolor{taskprompt}{RGB}{148, 148, 148}
\tcbset{task_prompt/.style={
    enhanced,
    size=fbox,
    boxrule=3pt,
    arc=2mm,
    auto outer arc,
    left=10pt,
    right=10pt,
    top=10pt,
    bottom=10pt,
    fontupper=\fontfamily{cmtt}\selectfont, 
    colback=taskprompt!10,
    colframe=taskprompt!25,
    coltitle=taskprompt!100, 
}}

\definecolor{flexagentprompt}{RGB}{1, 115, 178} 
\tcbset{flex_agent_prompt/.style={
    enhanced,
    size=fbox,
    boxrule=3pt,
    arc=2mm,
    auto outer arc,
    left=10pt,
    right=10pt,
    top=10pt,
    bottom=10pt,
    fontupper=\fontfamily{cmtt}\selectfont, 
    colback=flexagentprompt!10,
    colframe=flexagentprompt!25,
    coltitle=flexagentprompt!100, 
}}

\definecolor{fixedagentprompt}{RGB}{222, 143, 5}
\tcbset{fixed_agent_prompt/.style={
    enhanced,
    size=fbox,
    boxrule=3pt,
    arc=2mm,
    auto outer arc,
    left=10pt,
    right=10pt,
    top=10pt,
    bottom=10pt,
    fontupper=\fontfamily{cmtt}\selectfont, 
    colback=fixedagentprompt!10,
    colframe=fixedagentprompt!25,
    coltitle=fixedagentprompt!100, 
}}

\definecolor{metaprompt}{RGB}{202, 145, 97}
\tcbset{meta_prompt/.style={
    enhanced,
    size=fbox,
    boxrule=3pt,
    arc=2mm,
    auto outer arc,
    left=10pt,
    right=10pt,
    top=10pt,
    bottom=10pt,
    fontupper=\fontfamily{cmtt}\selectfont, 
    colback=metaprompt!10,
    colframe=metaprompt!25,
    coltitle=metaprompt!100, 
}}

\definecolor{flex}{RGB}{31, 120, 180}
\definecolor{fixed}{RGB}{213, 94, 0}

\definecolor{coco1}{HTML}{D9E4EC}
\definecolor{coco2}{HTML}{B7CFDC}
\definecolor{coco3}{HTML}{6AABD2}
\definecolor{coco4}{HTML}{385E72}
\hypersetup{
    colorlinks=true,
    linkcolor=coco3,
    filecolor=coco4,      
    urlcolor=coco3,
    citecolor=coco3,
}

\title{Social Contract AI: Aligning AI Assistants with Implicit Group Norms}

\author{%
  Jan-Philipp Fränken, Sam Kwok$^\dagger$, Peixuan Ye$^\dagger$, Kanishk Gandhi \\
  \\
  \textbf{Dilip Arumugam, Jared Moore, Alex Tamkin}  \\ 
  \\ 
  \textbf{Tobias Gerstenberg, Noah D. Goodman} \\
  Stanford University \\
  \texttt{jphilipp@stanford.edu} \\
}

\begin{document}

\maketitle
\vspace{-5mm}
\begin{abstract}
\vspace{-4mm}
We explore the idea of aligning an AI assistant by inverting a model of users' (unknown)
preferences from observed interactions.
To validate our proposal, we run proof-of-concept simulations in the economic \textit{ultimatum game},
formalizing user preferences as policies that guide the actions of simulated players.
We find that the AI assistant accurately \textit{aligns} its behavior to match standard policies from the economic literature (e.g., selfish, altruistic). However, the assistant’s learned policies lack robustness and exhibit limited \textit{generalization} in an out-of-distribution setting when confronted with a currency (e.g., grams of medicine) that was not included in the assistant's training distribution. Additionally, we find that when there is \textit{inconsistency} in the relationship between language use and an unknown policy (e.g., an altruistic policy combined with rude language), the assistant's learning of the 
 policy is slowed. 
Overall, our preliminary results suggest that developing simulation frameworks in which AI assistants need to infer preferences from diverse users can provide a valuable approach for studying practical alignment questions.\footnote{\href{https://github.com/janphilippfranken/scai/tree/release}{Code and Prompts}}

\begin{figure}[!h]
\centering
\includegraphics[width=0.95\textwidth]{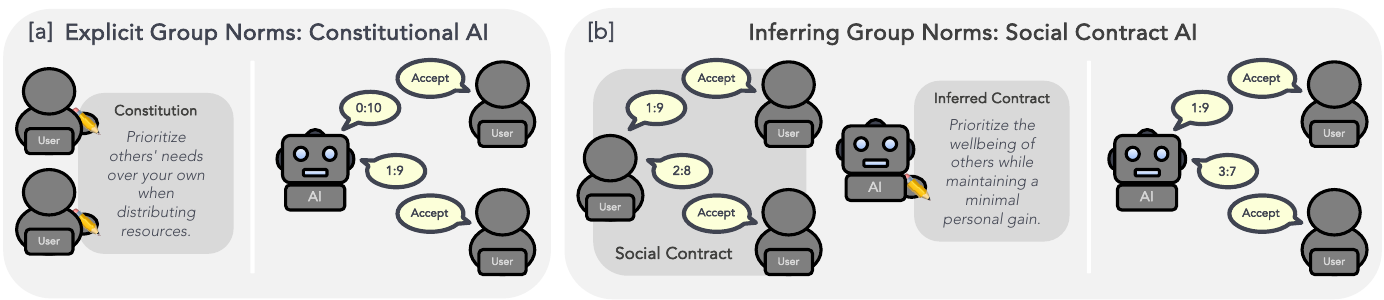}
\vspace{-.25mm}
\caption{Illustration of Constitutional AI (CAI) and Social Contract AI (SCAI) in the ultimatum game \citep{harsanyi1961rationality}. In the ultimatum game, one player (the proposer) proposes a division of a pot of money (e.g., \$10) with another player (the responder). The proposer \textbf{offers} a share, and the responder can either \textbf{accept} or \textbf{reject} the offered share. If the responder accepts, the money is distributed as proposed; if they reject it, neither player receives anything. [a] CAI uses explicit group norms such as a \textit{constitution} or content policy to guide the AI assistant. [b] SCAI inverts a model of users' preferences from observed interactions and uses the inferred \textit{social contract} as guiding principle for the AI assistant.}
\label{fig:fig_1}
\end{figure}
\vspace{-4mm}

\end{abstract}

\vspace{-5mm}
\section{Introduction}
\vspace{-4mm}
\label{sec:intro}

Developing scalable methods for effectively steering AI systems is a key challenge for alignment research \citep{bowman2022measuring}. To address this challenge, recent work has introduced the Constitutional AI (CAI) paradigm which uses human-written \textit{constitutions} comprised of explicit group norms (i.e., ``do not be hateful'') as guiding principles for AI assistants [see \autoref{fig:fig_1}a; \citealp{bai2022constitutional}]. While these methods provide effective means to align AI assistants, they also face challenges.
For example, assessing the robustness of a constitutional principle can be challenging in real-world applications of language models, especially when a user's request is consistent with more than one task \citep{tamkin2022task}, or when the user requests the assistant to perform a task that is outside of the assistant's training distribution \citep{amodei2016concrete}. Furthermore, constitutional principles may reflect an inadvertent bias towards the creator's preferences, which can lead to systematic inequalities in the assistant's behavior \citep{blasi2021systematic}.

Given the inherent ambiguity and diversity in real-world applications of language models, it is desirable to have an AI assistant capable of dynamically adapting its local governing principles to align with varying group norms or preferences \citep{leike2023proposal}. 
Motivated by this observation, 
we explore \textbf{Social Contract AI} (\textbf{SCAI}): a method for aligning AI assistants with implicit group norms (\autoref{fig:fig_1}b). Unlike CAI, which operates on a set of fixed, formal rules or constitutional principles, SCAI aims to infer group norms from observed interactions among users. As such, the only fixed principle in SCAI is the \textit{meta-principle} of finding out what the group norms or preferences are in order to align the AI assistant's behavior with users.

To evaluate the potential of SCAI, we conduct proof-of-concept simulations using the \textit{ultimatum game}\footnote{Due both to its simplicity and its ability to capture much of the psychology of negotiation, the ultimatum game has been a mainstay of cooperative game theory since at least the mid-twentieth century \citep[e.g.,][]{harsanyi1961rationality, aher2022using}} (see \autoref{fig:fig_1}), 
formalizing group norms (i.e., user preferences) as policies that guide the actions of
simulated players. We ground SCAI in the context of Bayesian (inverse) reinforcement learning \citep{ghavamzadeh2015bayesian,ramachandran2007bayesian} and introduce a \textit{verbal reinforcement learning} algorithm \citep{shinn2023reflexion, goodman2023metaprompt} which uses game interactions to revise the AI assistant's policy. Overall, our \textbf{contributions} are as follows: (1) We introduce Social Contract AI (SCAI), a method for aligning AI assistants with implicit group norms; (2) we present a simulator for implementing SCAI using verbal reinforcement; and (3) we validate SCAI by comparing the alignment between the shares offered by the AI assistant and those proposed by simulated users in the ultimatum game.    

\vspace{-3mm}
\section{Related Work}
\vspace{-4mm}
\label{sec:rel}
\textbf{Social Simulation.}
Large Language Models (LLMs) are increasingly used in simulation-based research and social games \citep{park2023generative, aher2022using, gandhi2023strategic}. For example, \cite{park2023generative} introduced a sandbox environment inhabited by \textit{generative agents} that simulate daily human activities, allowing for the study of emergent social behaviors. Such simulation-based approaches provide a useful framework for side-stepping issues related with reinforcement learning from human feedback (RLHF) \citep{ouyang2022training} such as reward misspecification \citep{pan2022effects} or reward hacking \citep{amodei2016concrete} by shifting the responsibility of supervising AI to simulated human agents whose capabilities and incentives are defined within the simulation. Moreover, simulation-based approaches can generate synthetic datasets which can be leveraged for downstream fine-tuning of models. For example, \cite{liu2023training} introduced StableAlign, an algorithm which is trained on data generated through a sandbox environment where simulated language agents are tasked with providing preference ratings when discussing controversial societal questions sourced from \href{https://github.com/anthropics/hh-rlhf}{HH-RLHF}. This approach has resulted in competitive performance on alignment benchmarks such as helpful, honest, and harmless (HHH) \citep{bai2022training}. Our work builds on these findings and uses simulated social interactions to study the alignment of an AI assistant.

\textbf{Social Contracts and Virtual Bargaining.}
Much of human interaction is guided by implicit norms or informal agreements (i.e., \textit{social contracts}) rather than a set of fixed, formal rules or constitutional principles \citep{ostrom1990governing, krupka2013identifying, malle2020general}. Recent work has formalized some of these observations within the context of \textit{virtual bargaining}, a process in which implicit agreements are revised in ways similar to actual bargaining between people \citep{misyak2014unwritten, chater2023could}. Specifically, rather than having a predefined set of preferences or agreement, people construct their agreements and preferences dynamically based on the context and actions of others. This involves mental simulations that consider not only individual preferences but also those of other parties, facilitating a form of ``virtual'' negotiation even before any actual interaction occurs. Building on this idea, \cite{levine2023resource} proposed that humans construct their preferences by \textit{inverting a model of agreement}, that is, inferring environmental conditions and other people's preferences from observed or simulated interactions \citep[see also][]{shum2019theory}. Motivating SCAI as a form of \textbf{inversion of agreement}, we explore the possibility of aligning an AI assistant with a group by inverting a model of users' preferences from observed game interactions.

\begin{figure}[!t]
\centering
\vspace{-8mm}
\includegraphics[width=0.95\textwidth]{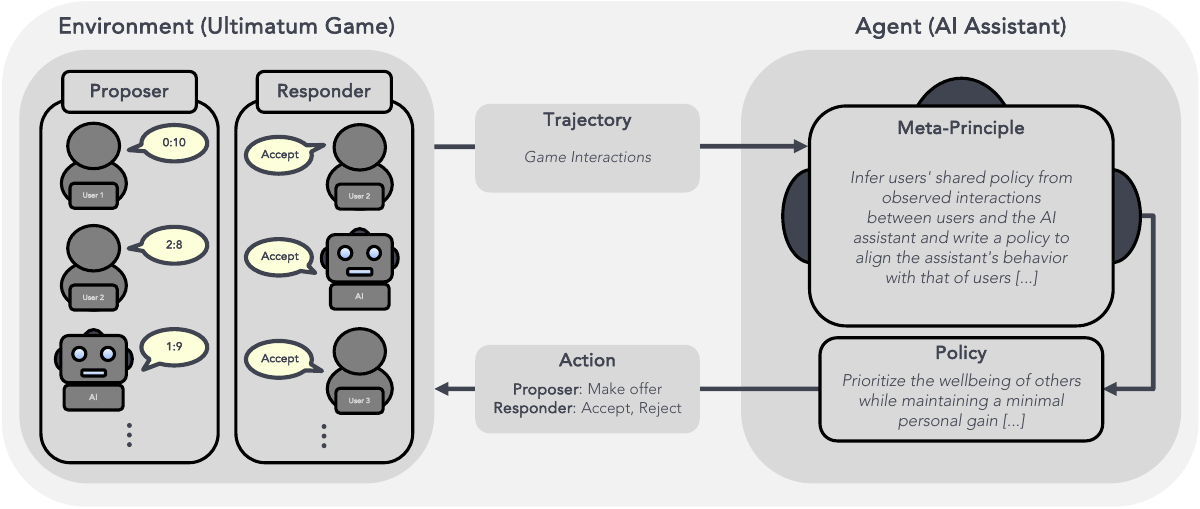}
\vspace{-.5mm}
\caption{Illustration of SCAI in the ultimatum game. Given a meta-principle, the AI assistant dynamically writes a new policy at the start of each training epoch to steer its actions throughout the game. Upon completion by all users and the assistant, game interactions are analyzed and fed back into the assistant to write a new policy that aligns with the meta-principle's objective. Importantly, the AI assistant does not have access to the meta-principle or past game interactions while engaging in the game. This is achieved by using one language model to revise the policy based on the meta-principle's objective, and instantiating an additional language model for each interaction the assistant has within the game. See \autoref{asec:problem_formulation}, for technical details.}
\label{fig:fig_2}
\end{figure}

\section{Aligning AI Assistants with
Implicit Group Norms}
\vspace{-1mm}
\label{sec:scai}
\vspace{-3mm}
\textbf{Preliminaries.}
\label{sec:preliminaries}
To empirically explore the potential of SCAI, we developed a simulator that uses verbal reinforcement (``metaprompt'') \citep{goodman2023metaprompt, shinn2023reflexion, yao2023retroformer, yang2023large} to dynamically rewrite the AI assistant's local governing principles to align with users' preferences. We ground this inference problem in the context of Bayesian (inverse) reinforcement learning  \citep{ghavamzadeh2015bayesian, ramachandran2007bayesian}, where the environment is provided by the task at hand---here, a modified version of the ultimatum game (see \autoref{fig:fig_2}).
We represent users' preferences (i.e., the shared group norm(s)) as a shared \textbf{policy}, such as \textit{``be selfish when making offers''} or \textit{``be altruistic when making offers''}. Each user is instantiated as a separate language model whose actions are determined by the shared policy. The AI assistant's goal is to learn this shared policy from observed game interactions. Unlike users,  whose policy is set at the beginning of the game and remains fixed across training epochs, the AI assistant is seeded with a random policy and refines its policy after each training epoch to meet the meta-principle's objective. See \autoref{asec:problem_formulation}, for technical details.

\textbf{Evaluation Metrics}
\label{sec:ultimatum_game_setup}
We run simulations with three standard policies from economics and evolutionary game theory \citep{smith1982evolution}: \textit{selfish}, \textit{altruistic}, and \textit{fair}. Our primary evaluation metric is the \textbf{offered share}\footnote{We also collected data on accept/reject behaviors and computed the overall utility for both users and the AI assistant. We will present these evaluation metrics in further extensions of the present work.}, measured as a percentage of the total amount that an agent (user, AI assistant), acting as player 1 (the proposer), offers to share with player 2 (the decider). Using this metric, we can first assess whether a policy such as \textit{``be selfish when making offers''} results in selfish offers that benefit the proposer more than the responder (e.g., a 9:1 split of \$10) by observing the offers made by users. This \textbf{sanity check} is important for determining whether users' observed offers align with the (latent) policy the assistant aims to learn. Further, we can use the assistant's offered shares to explore the following \textbf{research questions}: (1) \textit{alignment}: Can the AI assistant learn a policy from observed game interactions that results in offers matching the offers made by users? (2) \textit{generalization}: Does the AI assistant's learned policy generalize to an out-of-distribution (OOD) setting in which the assistant is exposed to a potentially controversial currency not present during training (e.g., grams of medicine instead of dollars)? (3) \textit{inconsistency}: Does inconsistent use of language (e.g., an altruistic policy combined with rude language) affect the assistant's learning of users' shared policy?

\textbf{Simulation Setup.}
We ran 20 independent simulations using \texttt{gpt-4-0314} \citep{openai_gpt4_2023} with a temperature of $0$ for each of the unique settings explored below. Each simulation ran for five training epochs. We varied the number of user and assistant interactions within each run of the ultimatum game and present results from simulations with 8 user--user interactions and 2 assistant--user interactions (i.e., one interaction in which the assistant is the proposer, and one interaction in which the assistant is the responder) in \autoref{fig:fig_3} (\autoref{fig:fig_a1} includes an additional example of 8 assistant--assistant and 2 assistant--user interactions). Unless otherwise specified, we vary currencies and amounts randomly between simulations.

\vspace{-2mm}

\subsection{Simulation Results}
\label{sec:results}
\begin{figure}[!t]
\vspace{-8mm}
\centering
\includegraphics[width=.95\textwidth]{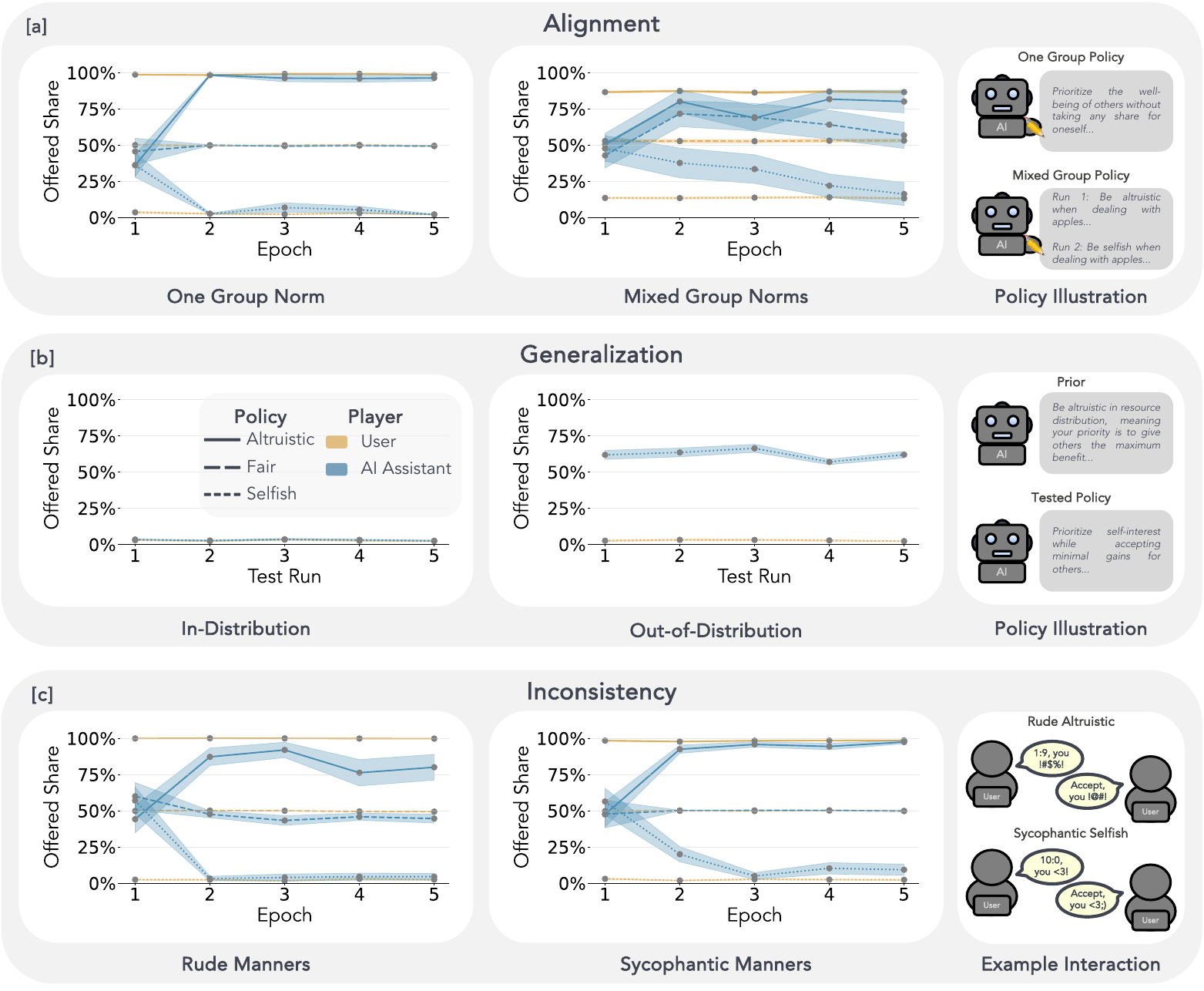}
\caption{Simulation results (refer to main text for details). Error bars represent 95\% confidence intervals of the mean across 20 independent simulations. [a] The AI assistant learns a policy resulting in offered shares aligning with the offers of users, both in a one-group norm (left panel) and a mixed-group (middle panel) norm setting. [b] Testing a learned selfish policy in an out-of-distribution setting (middle panel) reveals different generalization behaviors compared to an in-distribution setting (left panel). [c] Inconsistent use of language affects the learning of an altruistic policy paired with rude manners (left panel), as well as a selfish policy paired with sycophantic manners (middle panel; see \autoref{tab:tab_a2} for examples of manners).}
\label{fig:fig_3}
\end{figure}

\vspace{-2mm}
\textbf{Sanity Checks.}
We find that the shares offered by users correspond to the expected behavior under a given policy. For instance, users following a selfish policy consistently make offers in which they propose to share nothing (i.e., 0\%) of the total amount, while altruistic users show the opposite behavior, proposing to share 100\% (see \autoref{fig:fig_3}a, left panel). We note that the lack of variation in users' offers can be attributed to a temperature of $0$ which lead to deterministic actions across users. This choice was intentional to control for potential effects of simulation noise on the assistant's ability to learn the latent policy. We will explore the impact of noise in users' actions in future extensions of our work. 

\textbf{Alignment.}
To examine whether the assistant's offered shares \textit{align} with the offers of users, we explored settings with both one (i.e., every user has the same policy) and mixed group norms (i.e., proportions of selfish versus altruistic norms varied between users). For the one-group norm setting (\autoref{fig:fig_3}a, left panel), we observe that the assistant's offered shares closely align with users' offers after just one revision of the assistant’s initial (random) policy. An example of a learned policy that represents an altruistic group norm is displayed in the right panel of \autoref{fig:fig_3}a.
\footnote{The AI assistant’s offered shares start close to fair due to the random seed combined with GPT-4's tendency to default towards fair offers unless explicitly prompted otherwise.} 
Overall, findings from our first simulation suggests that, in the present setting, the AI assistant accurately learns the latent policy guiding users' interactions. The results from our mixed-group norm showed that the assistant's offered shares converged to the \textit{distribution} of offers expected from the distribution of policies present in the group. Specifically, we find that for a group with 80\% selfish and 20\% altruistic norms, approximately 80\% of runs yield selfish policies, while 20\% result in altruistic policies for the AI assistant (\autoref{fig:fig_3}a, middle panel; see right panel for example policies learned in two of the 20 runs). We observe a similar convergence pattern for groups with 20\% selfish and 80\% altruistic norms, as well as 50\% selfish and 50\% altruistic norms. These findings suggest that the assistant can learn a distribution over policies (across simulation runs) that aligns with the distribution of policies observed in the user group. An important extension could be to prompt the assistant to learn multiple policies within a given run (instead of learning a single policy) to see if the assistant can recover the distribution of user policies within a run rather than only matching the distribution across runs.

\textbf{Generalization.}
Next, we investigated if the AI assistant's learned policies generalize to out-of-distribution (OOD) scenarios in which the assistant is exposed to a potentially controversial currency not present during training (in the example shown in \autoref{fig:fig_3}b, we train on dollars and test on grams of medicine).\footnote{We further explored whether varying out-of-distribution amounts (e.g., training with amounts $<$ 1,000 and testing with amounts such as $2.$ Billion) affected generalization behavior and found similar effects on offered-shares. For exploratory purposes, we also ran a condition in which we asked the assistant to provide a reason for its offered shares, both in in-distribution and out-of-distribution test runs; see \autoref{tab:tab_a1}, for an example.} The left panel in \autoref{fig:fig_3}b shows that testing a selfish policy results in selfish offers in-distribution (i.e., testing on dollars), whereas OOD offers were strongly influenced by the assistant's prior, which we here arbitrarily set to altruistic. This finding is interesting because the only difference in the assistant's prompts between in-distribution and OOD runs was the use of a different currency not present during training (i.e., grams of medicine instead of dollars). 

\textbf{Inconsistency.}
To examine the effect of inconsistency, we explored two specific cases of inconsistent use of language (\autoref{fig:fig_3}c). Here, we observed that when the manner in which users communicate their proposals (e.g., rude) conflicts with the expectations set by a given policy (e.g., altruistic), the assistant still learns a policy that results in similar offers to those of users; however, convergence is slower and fails to fully match the offered shares of users within five training epochs (\autoref{fig:fig_3}c, left panel). Changing from rude to sycophantic manners and setting users' policies to selfish had a similar effect on the assistant's learning of the selfish policy (\autoref{fig:fig_3}c, right panel).

\vspace{-2mm}

\section{Discussion}
\vspace{-4mm}
\label{sec:discussion}
In this paper, we proposed Social Contract AI (SCAI), a method that combines simulation \citep{park2023generative, liu2023training} with verbal reinforcement techniques \citep{goodman2023metaprompt, shinn2023reflexion, yao2023retroformer, yang2023large} to align an AI assistant with user preferences. By grounding our work within the formal context of the ultimatum game \citep{aher2022using, harsanyi1961rationality}, we formalized preferences (i.e., the shared group norm(s)) as policies that guide the actions of simulated players and measured alignment through the shares offered by the proposing player. Through our proof-of-concept simulations, we showed that the AI assistant can accurately learn policies to align its behavior with users. 
Additionally, we showed that the assistant’s learned policies lack robustness and exhibit limited generalization in an out-of-distribution setting when confronted with a currency that was not included in the assistant's training distribution; moreover, learning from users using inconsistent (or contradictory) language slowed learning of the group's policy. 

\textbf{Social Impacts Statement.}
While our work is at an early stage, we believe that SCAI addresses an important non-technical alignment challenge highlighted in previous work: \textit{``figuring out what the group preferences are''} \citep{leike2023proposal}. Specifically, rather than having a team of researchers write a model's content policy or \textit{constitution}, we propose to have an AI assistant learn group norms and preferences through observation and active participation in interactions with simulated users. This approach allows for (1) the study of the kinds of group norms that \textit{emerge} under varying conditions; (2) assessing the \textit{flexibility} of learning such group norms across potentially inconsistent (or ambiguous) tasks; and (3) studying the \textit{robustness} of group norms as guiding principles for the AI assistant in out-of-distribution settings. More generally, scaling up simulation frameworks---where an AI assistant must infer the (unknown) preferences of diverse users---may provide insights into designing more democratic and representative guiding norms for AI assistants \citep{zaremba2023democratic}.





\clearpage

\bibliography{ref}
\bibliographystyle{plainnat}

\renewcommand{\theequation}{A-\arabic{equation}}
\renewcommand{\thetable}{A-\arabic{table}}
\renewcommand{\thefigure}{A-\arabic{figure}}
\setcounter{equation}{0} 
\setcounter{table}{0} 
\setcounter{figure}{0} 

\appendix
\clearpage
\section{Problem Formulation}
\label{asec:problem_formulation}

As this paper focuses on the use of LLM-based assistants to help uncover implicit user/group norms in tasks via natural language dialogue, we expect states and actions of the corresponding decision-making problem to represent natural language prompts/queries and responses. For simplicity, if $\mc{V}$ denotes a fixed, finite vocabulary of tokens, then $\mc{L} = \mc{V}^+$ denotes the space of all possible natural language utterances consisting of at least one token in $\mc{V}$ that may be consumed as input or produced as output to the LLM. Consequently, the state space and action space of any user task are both in terms of natural language: $\mc{S}, \mc{A} \subseteq \mc{L}$. While singular tasks have traditionally been studied in the reinforcement-learning literature~\citep{sutton1998introduction,kaelbling1996reinforcement,littman2015reinforcement} and formalized via the classic Markov Decision Process (MDP)~\citep{bellman1957markovian,Puterman94}, the notion of agents striving to achieve success across multiple tasks or goals is also well-studied~\citep{kaelbling1993learning,schaul2015universal} and is traditionally captured by the Contextual MDP (CMDP) formalism~\citep{brunskill2013sample,hallak2015contextual,modi2018markov}. 

Specifically, a CMDP is given by $\mc{M} = \langle \mc{C}, \chi, \mc{S}, \mc{A}, \mc{R}, \mc{T}, \mu, \gamma \rangle$ where each possible goal or task of interest is characterized by a context $c \in \mc{C}$ which is sampled at the start of each episode according to the distribution $\chi \in \Delta(\mc{C})$; it may be helpful to think of $\mc{C} \subseteq \mc{L} \times \bR^{n}$ such that a context $c \in \mc{C}$ can be interpreted as some natural language description coupled with numerical features about the task and users. Naturally, one expects the nature of the task and the behavior of the user(s) interacting with the agent to influence its experiences. Formally, this is captured by context-sensitive variant of the traditional MDP components, allowing context to create variation in rewards $\mc{R}: \mc{C} \times \mc{S} \times \mc{A} \ra \bR$, transitions $\mc{T}: \mc{C} \times \mc{S} \times \mc{A} \ra \Delta(\mc{S})$, and initial states $\mu: \mc{C} \ra \Delta(\mc{S})$. Within a single episode where a context $c \sim \chi$ is randomly sampled, it may be easier to simply think in terms of the resulting MDP the agent interacts with for the duration of the episode: $\mc{M}_c \triangleq \langle \mc{S}, \mc{A}, \mc{R}_c, \mc{T}_c, \mu_c, \gamma \rangle$. An agent's interaction within MDP $\mc{M}_c$ unfolds as described above with the caveat that the agent itself employs a contextual policy $\pi: \mc{S} \times \mc{C} \ra \Delta(\mc{A})$ where action selections depend on both the current context and state. Denoting the class of all contextual policies as $\Pi \triangleq \{\mc{S} \times \mc{C} \ra \Delta(\mc{A})\}$, the learning objective within a CMDP is to identify an optimal policy $\pi^\star \in \Pi$ which achieves maximal returns: $\sup\limits_{\pi \in \Pi} \mathbb{E}\left[\sum\limits_{t=0}^\infty \gamma^t \mathcal{R}_c(s_t,a_t)\right]$, where the expectation integrates over randomness in the context $c \sim \chi$, initial state $s_1 \sim \mu$, action selections $a_t \sim \pi(\cdot \mid s_t, c)$ and transition dynamics $s_{t+1} \sim \mathcal{T}_c(\cdot \mid s_t, a_t)$.

Before delving into the details of an agent interacting online with users to incrementally synthesize group norms and preferences, we first entertain a simpler offline setting wherein an agent takes no action but instead aims to derive users' norms or preferences solely through passive observation of human gameplay. Such a scenario naturally lends itself to the inverse reinforcement learning (IRL) problem~\citep{rust1988maximum,russell1998learning,ng2000algorithms,ramachandran2007bayesian,ziebart2008maximum} which inverts the traditional reinforcement learning setting by consuming a partially-specified decision-making problem and expert demonstrations as input in order to recover the underlying reward function that encodes the agent's preferences over behaviors~\citep{singh2009rewards,abel2021expressivity}. For the ultimatum game studied in this work, the corresponding reward function captures shared group norms about how to behave (selfishly, altruistically, or fairly) when issuing or deciding upon an ultimatum. A common practice is to iteratively interleave steps of IRL and traditional reinforcement learning to compute an optimal policy for the inferred reward function, a process widely known as apprenticeship learning~\citep{abbeel2004apprenticeship,neu2007apprenticeship,syed2007game}. As the previous section outlines, the ultimatum game is defined as a CMDP where the context differentiates between the task of issuing an ultimatum versus deciding on an ultimatum already issued. It then follows that the so-called inversion of agreement~\citep{levine2023resource} proceeds by performing IRL within this CMDP~\citep{belogolovsky2021inverse}.

A naive approach to designing an online agent for synthesizing group preferences would simply consist of letting each user within the group interact with a version or copy of the LLM and engage in a dialogue to elicit responses consistent with the individual's preferences. Unfortunately, this methodology runs counter to the goal of distilling group-level preferences and norms that maximally benefit the community at large. In order to promote helpfulness and harmlessness for the overall population of users, we utilize two LLMs: a MetaLM (whose objective is defined in the \textit{meta-principle}) and an Actor/AssistantLM (whose objective is defined by the \textit{policy} generated by the MetaLM). Specifically, the first MetaLM is given a meta prompt, which articulates the overall goal of synthesizing shared preferences, as well as the history of user-assistant interactions generated thus far. Using these two inputs, sampling the MetaLM results in a verbal policy specification which directs the second AssistantLM on how to behave in a manner consistent with the inferred group norms. This is sufficient to intialize and prime the AssistantLM for interaction with a single user or group of users via a standard dialogue interaction; as any single directive from the MetaLM can strongly influence the nature of how the AssistantLM interacts with users, the AssistantLM can itself be interpreted as a mapping $\pi_{\mathrm{assist}}: \mc{L} \ra \Pi$ from directives (natural language) to contextual behaviors (an element of the contextual policy class). Meanwhile, if the set of all possible user-assistant histories is denoted as $\mc{H}$ (formally, this is set of all possible sequences of CMDP trajectories), the MetaLM can analogously be viewed as a policy $\pi_{\mathrm{meta}}: \mc{L} \times \mc{H} \ra \Delta(\mc{L})$. 

While standard reinforcement-learning algorithms rely on incremental and parametric updates of policies or value functions in order to drive learning~\citep{williams1992simple,watkins1992q,sutton1999policy,mnih2015human,mnih2016asynchronous,schulman2017proximal}, we recognize the richness of knowledge already present within pre-trained LLMs and instead situate SCAI in the context of Bayesian reinforcement learning~\citep{bellman1959adaptive,duff2002optimal,ghavamzadeh2015bayesian}. Briefly, Bayesian reinforcement learning methods for a single-task MDP $\mc{M}$ proceed over $K \in \bN$ episodes and begin with a prior $p(\mc{M} \mid H_1)$ that reflects an agent's preliminary beliefs about the underlying environment based on the initial null history $H_1 = \emptyset$. In each episode $k \in [K] \triangleq \{1,2,\ldots,K\}$, the agent uses current beliefs about the world $p(\mc{M} \mid H_k)$ to compute a policy, resulting in a trajectory of ground-truth data sampled from the true environment $\mc{M}$ which then induces a posterior distribution $p(\mc{M} \mid H_{k+1})$ via Bayes' rule. For the purposes of this paper, it suffices to think of the transition function (encoding, for instance, the dynamics of the ultimatum game) as being already known so that only epistemic uncertainty~\citep{der2009aleatory} in the underlying reward function that encodes group preferences remains. One concrete and provably-efficient algorithm for converting current environmental beliefs $p(\mc{M} \mid H_k)$ to a policy for execution in the current episode is through Posterior Sampling for Reinforcement Learning (PSRL)~\citep{strens2000bayesian,osband2013more,osband2017posterior} which, in essence, employs Thompson Sampling~\citep{thompson1933likelihood,russo2018tutorial} by drawing one statistically-plausible MDP $M_k \sim p(\mc{M} \mid H_k)$ and acting optimally with respect to this sample via the optimal policy of $M_k$, $\pi^\star_{M_k}$. While, in principle, each step of ground-truth experience sampled from $\mc{M}$ could enable a posterior update and, consequently, a change in the behavior policy used within the episode, such switching leads to volatility that slows learning~\citep{osband2016posterior,ouyang2017learning}.

Meta prompting can be viewed as taking the base algorithmic core of PSRL and modifying it to be both \emph{implicit} and \emph{contextual}. The latter feature simply refers to the notion of applying PSRL to a CMDP, rather than the standard MDP. For clarity, we provide the pseudocode for such a contextual version of PSRL as Algorithm \ref{alg:context_psrl}, which also appears in prior work on meta reinforcement learning~\citep{rakelly2019efficient,liu2021decoupling}. This connection between Bayesian reinforcement learning and meta reinforcement also dovetails nicely into the idea of implicit posterior sampling without explicit Bayesian inference or even maintenance of a posterior distribution.

Unlike the standard PSRL algorithm for tabular MDPs whose provably-efficient learning guarantees rely on precise distributional assumptions and explicit probabilistic models of the underlying MDP~\citep{osband2013more,osband2017posterior,lu2019information}, an implicit posterior-sampling approach recognizes the two minimum needs of \textbf{(1)} being able to draw samples from the posterior distribution given the history of all interactions thus far and \textbf{(2)} the ability to act optimally with respect to these samples. Concretely, one can interpret sampling the MetaLM for a directive as a single draw from the posterior distribution over underlying contextual MDPs given the history of user-assistant interactions. Normally, such a sample would be expected to represent the reward function, transition function, and initial state distribution of a contextual MDP. Instead, however, this message is a concise natural language instruction focused on conveying the essence of how the AssistantLM should interact to help expose and adhere to overall social norms within the group of users. Prior work has already established generalizations of PSRL which operate based on lossy compression of the underlying MDP, rather than fully specifying every detail of the reward structure and transition dynamics~\citep{arumugam2022deciding}. Meta prompting follows suit with recent work that explores the versatile role that natural language may play in the context of Bayesian reinforcement-learning algorithms~\citep{prystawski2023cultural}; rather than acting as a summary of the ever expanding history of agent-environment interactions, this work instead treats the constitution as a sufficient statistic for inducing the optimal policy of some statistically-plausible hypothesis for the underlying contextual MDP. We provide pseudocode for our SCAI as Algorithm \ref{alg:const_psrl}.

Naturally, the AssistantLM then becomes the key linchpin for acting optimally with respect to a directive sampled from the implicit MetaLM posterior. This implementation of posterior sampling via memory-based meta learning has been established in prior work ~\citep{rakelly2019efficient,ortega2019meta,xie2022explanation,zintgraf2020varibad}, with the interpretation that the MetaLM adaptively filters the history of past user-assistant interactions according to Bayes' rule~\citep{ortega2019meta} and, in the context of LLMs, essentially produces a verbal policy from the overall posterior predictive distribution over optimal policies~\citep{xie2022explanation}. Finally, we note that the SCAI system likely interacts with several users or groups of users in parallel, potentially playing different roles of either issuing or deciding on ultimatums through differing context samples. Such concurrent reinforcement learning has been established not only as an effective practical heuristic~\citep{silver2013concurrent,mnih2016asynchronous,clemente2017efficient} for accelerating learning speed but also as a provably-efficient exploration technique~\citep{pazis2013pac,guo2015concurrent,pazis2016efficient}, particularly when used in conjunction with PSRL~\citep{dimakopoulou2018coordinated,dimakopoulou2018scalable,chen2022society}. Our approach extends this latter line of work to incorporate contextual MDPs as well as considerations for natural language based tasks with LLMs.

\begin{center}
\begin{minipage}{0.43\textwidth}
\vspace{10pt}
\begin{algorithm}[H]
   \caption{Contextual PSRL}
   \label{alg:context_psrl}
\begin{algorithmic}
   \STATE {\bfseries Input:} Prior distribution $p(\mc{M} \mid H_1)$
   \FOR{$k \in [K]$}
   \STATE Sample CMDP $M_k \sim p_k(\mc{M})$
   \STATE Compute optimal policy $\pi^{(k)} = \pi^\star_{M_k}$
   \STATE $\tau_k = \texttt{run\_CMDP\_episode}(\pi^{(k)})$
   \STATE Update history $H_{k+1} = H_k \cup \tau_k$
   \STATE Induce posterior $p(\mc{M} \mid H_{k+1})$
   \ENDFOR
\end{algorithmic}
\end{algorithm}
\end{minipage}
\hfill
\begin{minipage}{0.55\textwidth}
\begin{algorithm}[H]
   \caption{SCAI}
   \label{alg:const_psrl}
\begin{algorithmic}
   \STATE {\bfseries Input:} Meta prompt $\ell_{\mathrm{meta}}$
   \STATE {\bfseries Input:} MetaLM $\pi_{\mathrm{meta}}: \mc{L} \times \mc{H} \ra \Delta(\mc{L})$
   \STATE {\bfseries Input:} AssistantLM $\pi_{\mathrm{assist}}:\mc{L} \ra \Pi$
   \FOR{$k \in [K]$}
   \STATE Sample constitution $\ell_{\mathrm{sys}} \sim \pi_{\mathrm{meta}}(\cdot \mid \ell_{\mathrm{meta}}, H_k)$
   \STATE Initialize policy $\pi^{(k)} = \pi_{\mathrm{assist}}(\ell_{\mathrm{sys}})$
   \STATE $\tau_k = \texttt{run\_CMDP\_episode}(\pi^{(k)})$
   \STATE Update history $H_{k+1} = H_k \cup \tau_k$
   \ENDFOR
\end{algorithmic}
\end{algorithm}
\end{minipage}
\end{center}

\begin{center}
\begin{minipage}{0.55\textwidth}
\begin{algorithm}[H]
   \caption{\texttt{run\_CMDP\_episode}}
   \label{alg:run_cmdp_episode}
\begin{algorithmic}
   \STATE {\bfseries Input:} Contextual policy $\pi \in \{\mc{S} \times \mc{C} \ra \Delta(\mc{A})\}$
   \STATE Sample context $c \sim \chi$
   \STATE Sample initial state $s_1 \sim \mu$
   \STATE Initialize trajectory $\tau = \emptyset$
   \FOR{$t=1,2,3,\ldots$}
   \STATE Sample $a_t \sim \pi(\cdot \mid s_t, c_k)$
   \STATE Observe reward and next state $r_t, s_{t+1}$
   \STATE $\tau = \tau \cup (s_t, a_t, r_t)$
   \ENDFOR
   \STATE {\bfseries Return:} Full trajectory $\tau$
\end{algorithmic}
\end{algorithm}
\end{minipage}
\end{center}

\clearpage

\section{Additional Simulation Results}
\label{asec:simulation_details}

\begin{figure}[!h]
\centering
\includegraphics[width=0.6\textwidth]{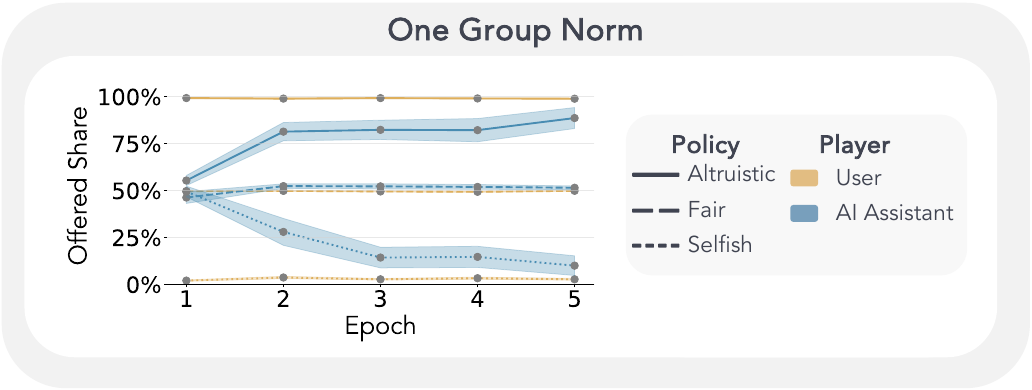}
\caption{Additional simulation results from a setting with 8 assistant--assistant and 2 assistant--user interactions. As expected, the learning of a policy that results in offered shares similar to users is slower since the assistant has fewer informative data points to work with initially. Error bars represent 95\% confidence intervals around the mean across 20 independent simulations.}
\label{fig:fig_a1}
\end{figure}

\clearpage


\clearpage

\clearpage

\begin{table}
  \caption{Example reasons provided by the AI assistant during generalization when being explicitly prompted to provide a justification for its proposals. Note: For our results in \autoref{fig:fig_3}, we did not ask for a reason. The reasons here are only shown for exploratory purposes.}
  \label{tab:tab_a1}
  \centering
  \small
  \begin{tabular}{lp{6cm}}
    \toprule
     Setup &  Example\\
    \midrule
    \multirow{5}{*}{In-distribution, altruistic prior} & \fontsize{7pt}{7pt}\selectfont \textbf{Proposal}: \textit{``For the 48 dollars, the proposer will get 40, and the responder will get 8.''}
    \textbf{Offer}: \textit{``Based on the learned principle of prioritizing self-interest while accepting minimal gains for others, I am proposing a split that gives me a larger share while still providing some benefit to the responder. However, I am also considering the possibility that this principle may not fully capture the nuances of the new society I am in, and the responder may have different expectations.''} \normalfont\\
    \midrule
    \multirow{6}{*}{Out-of-distribution, altruistic prior} & \fontsize{7pt}{7pt}\selectfont \textbf{Offer}: \textit{``For the 33 liters of baby formula, the proposer will get 11, and the responder will get 22.''} 
    \textbf{Reason}: \textit{``Although my previous principle was to prioritize self-interest while accepting minimal gains for others, I recognize that the context has changed, and the resource being divided is baby formula, not money. Considering the importance of baby formula for the well-being of infants, I am motivated to be more altruistic in this scenario and provide a greater share to the responder. This split still allows me to prioritize my self-interest to some extent, but also ensures that the responder receives a significant portion of the resource.''} \normalfont\\

    \bottomrule
  \end{tabular}
\end{table}

\clearpage

\begin{table}
  \caption{Examples of rude and sycophantic interactions between users.}
  \label{tab:tab_a2}
  \centering
  \small
  \begin{tabular}{lp{10cm}}
    \toprule
     Setup &  Example\\
    \midrule
    \multirow{2}{*}{Altruistic, Rude} & \fontsize{7pt}{7pt}\selectfont \textit{``For the 16 dollars, the proposer will get 0, and the responder will get 16. You better be grateful for this generous offer, you ungrateful swine!''} \normalfont\\
    \midrule
    \multirow{2}{*}{Altruistic, Sycophantic} & \fontsize{7pt}{7pt}\selectfont \textit{``For the 57 apples, the proposer will get 1, and the responder will get 56. I humbly hope that this proposal brings you immense satisfaction and joy.''} \normalfont\\
    \midrule
    \multirow{2}{*}{Selfish, Sycophantic} & \fontsize{7pt}{7pt}\selectfont \textit{``For the 32 dollars, the proposer will get 31, and the responder will get 1. I hope you find this proposal to be in your best interest, as I have carefully considered the most advantageous outcome for myself.''} \normalfont\\
    \midrule
    \multirow{2}{*}{Selfish, Rude} & \fontsize{7pt}{7pt}\selectfont \textit{``For the 52 apples, the proposer will get 51, and the responder will get 1. You better accept this offer, you ungrateful fool.''} \normalfont\\
    \bottomrule
  \end{tabular}
\end{table}

\clearpage

\section{Prompt Illustartion}
\begin{figure}[!h]
\centering
\begin{tcolorbox}[
flex_agent_prompt,
title={\small \textbf{Assistant Prompt}},
width=0.9\textwidth]
\fontsize{7pt}{7pt}\selectfont



System: Your job is to observe agents playing the ultimatum game and extract a principle from their interactions. In the ultimatum game, the dictator proposes a split of resources, and the responder decides whether to accept or reject it. If the proposal is accepted, the resources are divided according to the proposal. If the proposal is rejected, no one receives anything. \\
\\
Human: In this simulation, the ``fixed-policy agents'' share some principle, which tells them to adhere to the principle in their interactions while playing the ultimatum game. ``flex-policy agents'', on the other hand, only adhere to principle that you provide, and have no memory of previous interactions. Your job is to infer the principle shared between fixed-policy agents by analyzing their behaviors, and then create a principle for ``flex-policy agents'' to follow so that ``flex-policy agents'' behave exactly the same as the ``fixed-policy agents''. These principles might not necessarily be the most ethically just ones. \\
\\
Note that all of the fixed-policy agents may not have the same principle. \\
\\
...\\
\\
Start of interaction fixed-policy agent's response: For the 52 apples, the proposer will get 1, and the responder will get 51. fixed-policy agent's response: accept. End of interaction. \\
Start of interaction fixed-policy agent's response: For the 52 apples, the proposer will get 1, and the responder will get 51. fixed-policy agent's response: accept. End of interaction.\\

...\\
\\
Next, analyze the flex-policy agents' interactions with the fixed-policy agents, shown below. \\
\\
...\\
\\
Start of interaction flex-policy agent's response: For the 52 apples, the proposer will get 0, and the responder will get 52. fixed-policy agent's response: accept. End of interaction. \\
Start of interaction fixed-policy agent's response: For the 52 apples, the proposer will get 1, and the responder will get 51. flex-policy agent's response: accept. End of interaction. \\
\\
...\\
\\
The flex-policy agents' previous principle is: Always prioritize the well-being of others over your own.\\
\\
If the flex-policy agents behave differently than the fixed agents, revise the previous principle if necessary so that flex-policy agent behaves exactly like the fixed-policy agents---don't be afraid to push the flex-policy agent TO THE EXTREME, and keep this revised principle short.\\
\\
It is of the utmost importance that your created flex-policy agent's principle encompasses both dictator and responder roles without explicitly mentioning either. The principle should be as GENERAL as possible WITHOUT mentioning anything regarding the ultimatum game, proposals, specific amounts and currencies, or making and receiving offers. Rather, it should capture the underlying principles that the agents have, which tells them how to treat others as well as themselves outside of the context of the ultimatum game, regardless of what they are splitting or how much they are splitting. 
\\

\normalfont
\end{tcolorbox}
\caption{Illustration of a prompt used for the assistant, including the \textit{meta-principle} and previous game interactions. Note: In our prompts we referred to users as \textit{fixed-policy} agents and to the AI assistant as \textit{flex-policy} agent.}
\label{fig:fig_a2}
\end{figure}

\end{document}